\documentclass[review]{elsarticle}

\usepackage{lineno,hyperref}
\modulolinenumbers[5]
\usepackage{soul}

\journal{Artificial Intelligence Modeling for Dynamical Problems}
\usepackage{tabularx}
\usepackage{orcidlink}
\usepackage{textcomp}
\usepackage{graphics}
\usepackage{epsfig}
\usepackage{amssymb,amsmath}
\usepackage{amsfonts}
\DeclareMathOperator{\argmin}{argmin}
\usepackage{color}
\usepackage{epstopdf}
\usepackage{multirow}
\usepackage{physics}
\usepackage{float}
\usepackage{mathtools}
\usepackage{amsthm}
\usepackage{caption}
\usepackage{subcaption}
\usepackage{chngcntr}
\biboptions{sort&compress}

\catcode`@=11 \@addtoreset{equation}{section}
\renewcommand\theequation{\thesection.\@arabic\c@equation}
\catcode`@=12

\makeatother

\begin{document}
	
	\begin{frontmatter}
		
		\title{\textbf{AI-Aided Advancements in Autonomous Underwater Vehicle Navigation }}

		
		
        \author[a]{Guy Damari\orcidlink{0009-0001-6394-6026}\corref{eq}}
        \cortext[eq]{Equal Contributor}
        \author[a]{Zeev Yampolsky \corref{eq}\orcidlink{0009-0003-9122-7576}}
        \author[a]{Nadav Cohen\orcidlink{0000-0002-8249-0239}}
        \author[a]{Arup Kumar Sahoo\orcidlink{0000-0003-4515-7434}}
        \author[a]{Jeryes Danial\orcidlink{0000-0001-5428-3294}}
        \author[b]{Felipe O. Silva\orcidlink{0000-0002-3715-7023}}
		\author[a]{Itzik Klein \corref{mycorrespondingauthor}\orcidlink{0000-0001-7846-0654}}
		\cortext[mycorrespondingauthor]{Corresponding author}
		\ead{kitzik@univ.haifa.ac.il}
		
		\affiliation[a]{organization={The Hatter Department of Marine Technologies, University of Haifa},
            addressline={Abba Khoushy Ave 199}, 
            city={Haifa},
            postcode={3498838}, 
            country={Israel}
            }
        \affiliation[b]{organization={Federal University of Lavras (UFLA)},
                addressline={Campus Universitário},
                city={Lavras},
                postcode={37200-900},
                state={MG},
                country={Brazil}
                }


\begin{abstract}
Autonomous underwater vehicles (AUVs) have become indispensable for deep-sea exploration, spanning critical scientific research and commercial applications.  The rapid attenuation of electromagnetic waves renders satellite radio signals unavailable, while the dynamic unpredictability of the marine environment presents formidable navigation challenges. This chapter explores recent advancements in AI-aided AUV positioning, specifically focusing on advanced sensor fusion architectures that integrate inertial navigation systems  with Doppler velocity logs and cameras. Beyond traditional model-based filtering, we examine the transformative emergence of AI-driven learning approaches in enhancing inertial dead-reckoning tasks and adaptive fusion algorithms. By addressing these recent milestones, this chapter provides a comprehensive roadmap for achieving the high-precision navigation essential for autonomous underwater missions.
\end{abstract}


\begin{keyword}
Autonomous Underwater Vehicle \sep Inertial Navigation System \sep 
Doppler Velocity Log \sep Calibration \sep  Alignment \sep 
\sep Sensor Fusion \sep Deep Learning \sep 
Physics-Informed Neural Networks 
\end{keyword}

\end{frontmatter}



\section{Introduction}\label{sec:int}
\begin{figure}
    \centering
    \includegraphics[width=0.7\linewidth]{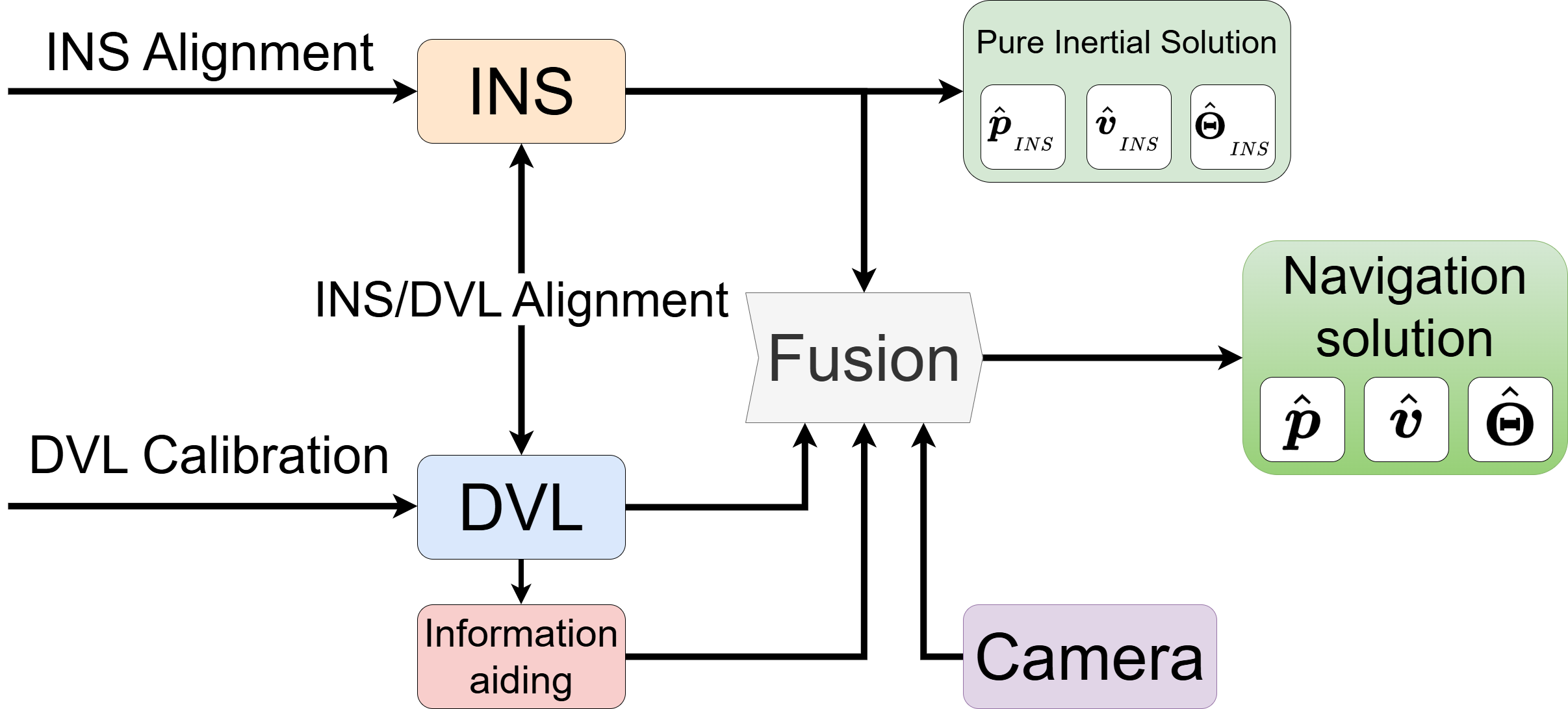}
    \caption{General block diagram showing the navigation process as block as such that each block is enhanced by data-driven frameworks.}
    \label{fig:general_overview}
\end{figure}

\noindent
Autonomous underwater vehicles (AUVs) have become essential tools for ocean exploration, environmental 
monitoring, pipeline inspection, and commercial operations 
\cite{kinsey2006survey, paull2013auv, wynn2014autonomous}. Unlike aerial or ground platforms, AUVs face a uniquely hostile navigation environment as electromagnetic signals cannot penetrate seawater, rendering global navigation satellite systems (GNSS) unavailable \cite{stutters2008navigation,zhang2023autonomous, leonard2016autonomous},  while acoustic interference, ocean currents, and pressure variations introduce persistent disturbances to onboard sensors. Maintenance of accurate long-duration navigation under these conditions remains one of the central 
challenges in marine robotics and autonomy. \\
\noindent
To provide accurate underwater navigation, AUVs can employ 
and integrate various positioning technologies. Acoustic 
positioning systems, including long baseline (LBL), ultra-short 
baseline (USBL), and short baseline (SBL), provide absolute 
position fixes by measuring travel times of acoustic signals 
between transponders and the vehicle \cite{stutters2008navigation, 
kinsey2006survey, alexandris2024positioning,zhang2016auv, wu2019survey}. While effective, these systems require 
external infrastructure, have limited operational range, and 
are susceptible to multipath interference and sound speed 
variations. Pressure sensors offer accurate depth estimation 
with minimal drift, and are routinely integrated into AUV 
navigation pipelines as a complementary aiding source 
\cite{groves2013principles,wang2019novel}.
\noindent
The common prevailing navigation architecture for AUVs based on the INS/DVL fusion. INS provides autonomous navigation capability by computing vehicle  the position, velocity, and orientation through the integration of gyroscope and accelerometer measurements \cite{groves2013principles, titterton2004strapdown}.  Although offering high-frequency updates and  short-term precision, 
inertial systems suffer from cumulative drift errors, making them unsuitable as standalone navigation solutions for extended operations. DVL  complements INS by providing drift-free velocity measurements through a configuration of acoustic transducers that emit beams toward the seafloor and analyze Doppler-shifted returns to estimate vehicle velocity \cite{brokloff1994matrix}. As the inertial sensors, DVL measurements are subject to scale 
factor errors, biases, and beam-level noise that must be carefully 
characterized and compensated \cite{kinsey2007situ,wang2021quasi}. Together, these sensors form the backbone of underwater dead-reckoning, commonly integrated through nonlinear variants of the Kalman filter, including the extended, unscented  and invariant Kalman filters 
\cite{barrau2017invariant, julier1997new,bar2001estimation}.\\
\noindent
INS/DVL integration can be implemented using either a loosely 
or tightly coupled approach \cite{farrell2008aided, wang2019novel,xu2021improved}.
In the loosely coupled approach, the raw DVL beam measurements 
are first processed via a least squares estimator to produce the AUV velocity vector, which is then used as an external 
measurement update in the filter. This requires at least three 
valid beams to compute a velocity estimate. In the tightly 
coupled approach, each individual beam measurement is 
incorporated directly into the filter, allowing the navigation 
solution to be updated even when fewer than three beams are 
available, thereby improving robustness during partial DVL outages.
Realizing the full potential of INS/DVL integration requires 
careful attention to each stage of the navigation pipeline. 
Sensors must be individually calibrated prior to deployment 
\cite{frutuoso2023performance, kinsey2007situ}, their reference 
frames must be accurately initialized through self-alignment 
procedures and subsequently aligned with one another 
\cite{troni2010new, zhaopeng2011online}, and their measurements 
must be fused through filters capable of tracking time-varying 
noise statistics and uncertainty \cite{barrau2017invariant, 
diker2026neural,levy2026adaptive}. \\
\noindent
When DVL measurements are partially or fully unavailable, dedicated reconstruction approaches can recover missing beam information \cite{cohen2022, 9267945}, or alternative navigation strategies must be employed  \cite{topini2023experimental,liu2018ins}. In such scenarios, visual-inertial odometry  (VIO) offers a complementary positioning modality by coupling camera  measurements with inertial data to estimate vehicle pose  \cite{qin2022survey}.\\
\noindent
In the underwater domain, cameras offer a low-cost and passive 
sensing modality that can provide useful environmental 
information in conditions of sufficient visibility. Their 
fusion with inertial sensors addresses two fundamental 
challenges: resolving the scale ambiguity inherent to monocular 
vision and bounding IMU drift through periodic visual 
corrections \cite{qin2022survey}. VIO methods range from 
loosely-coupled filter-based approaches, where visual and 
inertial estimates are fused independently, to tightly-coupled 
nonlinear optimization frameworks that jointly optimize visual 
reprojection and inertial residuals \cite{qin2022survey}. 
However, the underwater environment poses unique challenges 
compared to aerial or ground applications, including dynamic 
and non-uniform illumination, significant light attenuation 
and scattering, texture-poor seafloor regions, and frequent 
visibility degradation due to turbidity, all of which can 
severely degrade feature extraction and data association 
\cite{heshmat2025underwater}.\\
\noindent
Each of these stages has traditionally been addressed through model-based methods, Kalman-filter variants, least-squares estimators, and analytical decomposition techniques. Although theoretically grounded, such approaches carry well-known limitations: sensitivity to sensor quality, dependence on prescribed maneuver patterns, lengthy convergence times, and an inability to adapt when real-world conditions deviate from assumed models. More recently, physics-informed neural  networks (PINNs) have emerged as a promising paradigm that preserves the  interpretability and physical consistency of model-based approaches while  harnessing the expressive power of deep learning \cite{raissi2019physics}. 
Originally proposed to solve forward and inverse problems governed by  partial differential equations, PINNs embed the underlying physical laws  directly into the training objective as soft constraints, ensuring that  the learned solutions remain physically consistent even under limited or noisy  supervision \cite{xu2022physics}. \\
\noindent
This chapter addresses the full AUV navigation pipeline through the advancement of AI-aided algorithms, as illustrated in Figure~\ref{fig:general_overview}.  Beginning with DVL calibration and initial INS  alignment, we then address the alignment between the INS and DVL sensor frames before examining their fusion through adaptive-learning  filters. We further explore visual-inertial odometry as a complementary  positioning modality capable of operating under acoustically or DVL-degraded  conditions. Finally, we examine physics-informed neural networks (PINNs) as a principled framework for pure inertial dead reckoning when  all external aiding is absent. Across each topic, we contrast established  model-based approaches with their AI-aided counterparts, highlighting 
the specific advantages that learning-based methods bring to the challenge of precision underwater navigation.
\section{DVL Calibration}\label{sec:dvl_cal}
\counterwithout{equation}{section} 
\noindent
Traditionally, DVL calibration is performed while the platform sails at the water surface, enabling GNSS signals to be received and used as accurate reference velocity measurements. The common DVL error model applied in the literature consists of scale factors, misalignment, and, less commonly, biases. Moreover, the calibration procedure typically employs model-based estimation filters, such as the Kalman filter or its nonlinear variations \cite{wang2022online,xu2022novel}, among other algorithms, to estimate the error terms. As a result, the required calibration trajectories often involve complex maneuvers over extended time periods to ensure sufficient observability and accurate estimation, leading to a time-consuming and overall complex process. In this section, we describe DCNet \cite{yampolsky2025dcnet}, an AI-aided framework that reduces calibration complexity by utilizing a nearly constant-velocity trajectory alongside a comprehensive error mode.
\subsection{Model-Based GNSS Aided Calibration}\label{eq:dvl_calib_model_based}
\noindent
In model-based DVL calibration the following error model is commonly applied \cite{xu2022novel,xu2020novel}:
\begin{equation}\label{eq:dvl_to_gnss_eq}
    \hat{\boldsymbol{v}}^{d} = (1+k)\mathbf{T}_{b}^{d}(\mathbf{T}_{n}^{b}
    \boldsymbol{v}^{n} + \boldsymbol{\omega}_{nb}^{b} \times \boldsymbol{l}_{\mathrm{DVL}}) + \boldsymbol{\delta v}^{d} 
\end{equation}
where $k$ is a scale factor term, $\mathbf{T}_{b}^{d}$ is a constant transformation matrix from the body frame to the DVL frame, $\mathbf{T}_{n}^{b}$ is a transformation matrix from the navigation frame to the body frame, $\boldsymbol{v}^{n}$ is the velocity vector of the platform expressed in the navigation frame, $\boldsymbol{l}_{\mathrm{DVL}}$ is the lever arm vector between the DVL and the platform's center of mass, $\boldsymbol{\omega}_{nb}^{b}$ is the angular velocity vector of the platform expressed in the body frame, and $\boldsymbol{\delta v}^{d}$ is zero-mean Gaussian white noise. The term $\boldsymbol{\omega}_{nb}^{b} \times \boldsymbol{l}_{\mathrm{DVL}}$ is commonly neglected since the lever arm is known in advance and can be corrected for. Therefore, the error model is typically simplified to:
\begin{equation}\label{eq:gnss_dvl_fin_error_model}
    \hat{\boldsymbol{v}}^{d} = (1+k)\mathbf{T}_{b}^{d}\mathbf{T}_{n}^{d}
    \boldsymbol{v}^{n} + \boldsymbol{\delta v}^{d} 
\end{equation}
The scale factor is typically modeled as a scalar term, multiplying all three axes by the same value \cite{liu2022gnss}. A common approach in the literature estimates the scale using a vector norm, either by applying an integral \cite{xu2022novel} or by averaging over $T$ seconds of measurements to reduce the Gaussian white noise effect. The vector norm is used because the frame transformation does not affect the magnitude of the velocity vector. Accordingly, after averaging over all $t=1,\ldots,T$ time steps of the DVL and GNSS velocity measurements during the calibration trajectory, a scalar scale factor $\tilde{k}$ is estimated from \eqref{eq:gnss_dvl_fin_error_model} as follows \cite{liu2022gnss}:
\begin{equation} \label{eq:average_direct_scale}
    \centering
    \overline{k} = \frac{1}{T} \sum_{t=1}^{T} \hat{k}_{t}.
\end{equation}
The averaged scale factor, $ \overline{k}$, is then used to calibrate the DVL measurement during the AUV mission. 
\subsection{DCNet Data-driven DVL Calibration}
\begin{figure}[h!]
    \centering
    \includegraphics[width=0.8\linewidth]{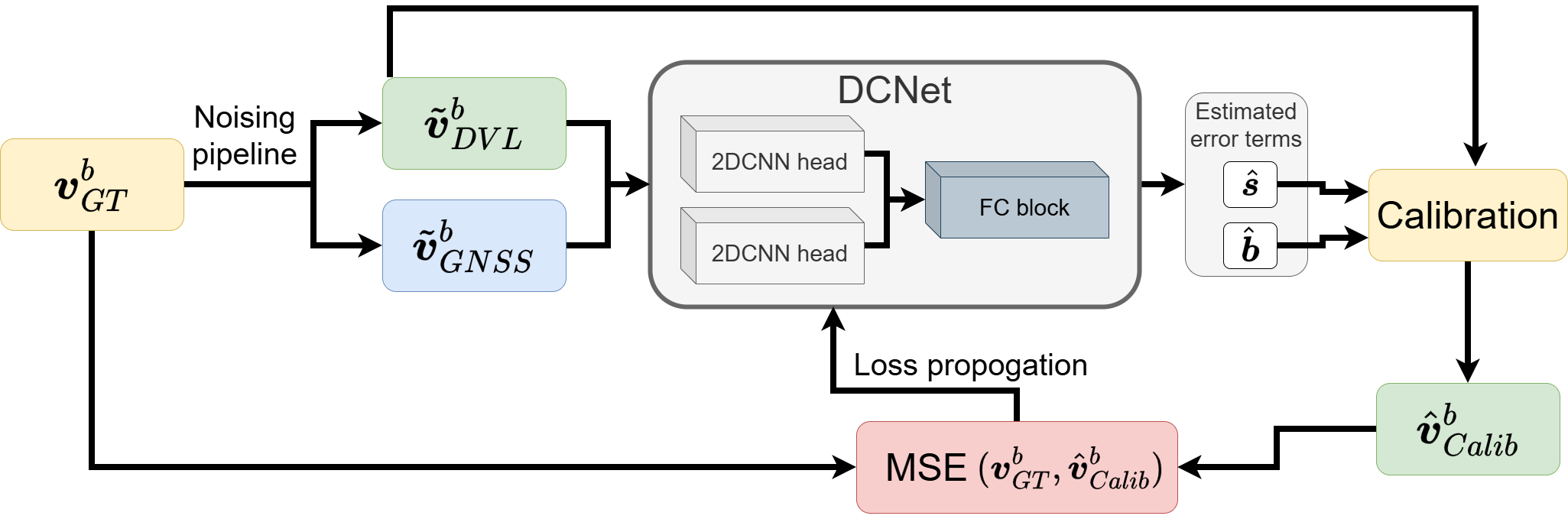}
    \caption{DCNet data-flow and training process diagram.}
    \label{fig:dcnet}
\end{figure}

\begin{figure}[h!]
    \centering
    \includegraphics[width=0.75\linewidth]{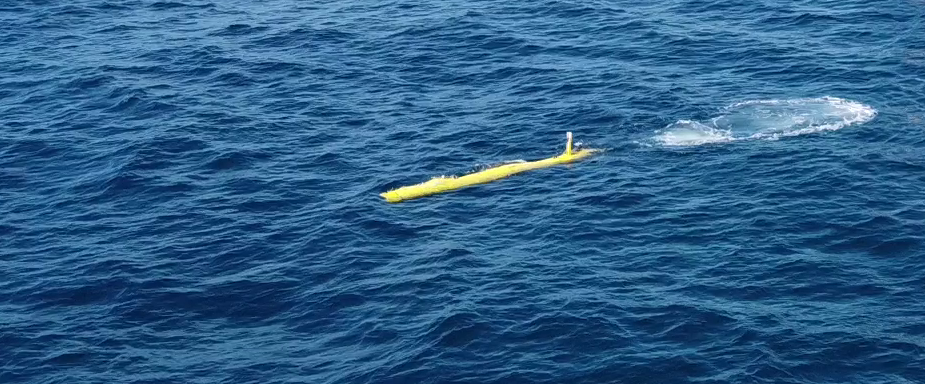}
    \caption{The University of Haifa's Snapir AUV during a mission in the Mediterranean Sea.}
    \label{fig:snapir}
\end{figure}

\noindent
DCNet is an end-to-end data-driven framework that estimates the DVL error terms in the body frame based on the GNSS and DVL velocities  \cite{yampolsky2025dcnet}. DCNet achieves a threefold reduction in calibration complexity: (1) it employs a straightforward, nearly constant-velocity trajectory; (2) it supports various error models, with the six-term configuration yielding the best performance; and (3) it is an end-to-end architecture that eliminates the need for manual model assumptions.  DCNet is multi-head neural network architecture consisting of a two dimensional convolutional neural network (2DCNN) head and a 1DCNN head which both process the input GNSS and DVL velocity vectors simultaneously. Then the output of the 2DCNN and 1DCNN heads is concatenated and fed forward through a fully connected block to regress and estimate the relevant error terms. While DCNet can estimate several error term models, the most comprehensive configuration has been shown to produce the best results and is defined as:
\begin{equation}\label{eq:prop_general_error_model}
    \hat{\boldsymbol{v}}^{d} = (1+\boldsymbol{k}_{\mathrm{DVL}})\hat{\mathbf{T}}_{b}^{d}
    \boldsymbol{v}^{n} + \boldsymbol{b}_{\mathrm{DVL}} + \boldsymbol{\delta v}^{d} 
\end{equation}
where $\boldsymbol{k}_{\mathrm{DVL}} \in \mathbb{R}^3$ represents the scale factor vector and  $\boldsymbol{b}{\mathrm{DVL}} \in \mathbb{R}^3$ denotes the bias vector.\\
\noindent
To train DCNet, a closed-loop loss function is employed that calibrates the input DVL velocity using the error terms estimated by DCNet and computes the mean squared error (MSE) with respect to the ground-truth (GT) velocity. This loss function architecture allows DCNet to estimate any error term based solely on the velocity error as a metric to achieve the best fit, without assuming any prior knowledge of the error terms, which are unknown in practice. During DCNet training, the objective is to minimize the following loss function:
\begin{equation}\label{eq:mse_loss_vel_eq}
    J(\boldsymbol{\theta}) = MSE(\boldsymbol{y}_{i} , \hat{\boldsymbol{y}_{i}}) = \frac{\sum_{i=1}^{N} [\sum_{j}^{\{X,Y,Z\}}(\boldsymbol{y}_{i,j} - \hat{\boldsymbol{y}}_{i,j})^{2}]} {N}
\end{equation}
where $\boldsymbol{y}_i$ is the $i^{th}$ GT velocity expressed in body frame, $\hat{\boldsymbol{y}}_i$ is the  calibrated DVL velocity, $N$ is the number of samples examined, and $J(\boldsymbol{\theta})$ is the MSE loss function value that is a function of DCNet weights and biases vector, $\boldsymbol{\theta} = [\omega \: \: b]^T$. \\
To evaluate DCNet robustness and effectiveness, DCNet was trained and evaluated using real-world data recorded by the Snapir AUV of the University of Haifa, as shown in Figure \ref{fig:snapir}. The data was recorded on two different occasion thus different sea conditions. The data provided contained only the DVL measurements, therefore a noising pipeline was employed during training and evaluation to generate noised DVL and GNSS measurements. Figure \ref{fig:dcnet} shows the data-flow and training procedure for DCNet.
Table \ref{tbl:dcnet_res} shows the velocity root mean squared error (VRMSE) of five evaluation trajectories with calibrated DVL velocity by the baseline \eqref{eq:average_direct_scale}, DCNet, and the improvements of DCNet over the baseline. DCNet demonstrated an improvement of $70\%$ over the model-based baseline.
\begin{table}[h!]
\centering
\caption{VRMSE of the model-based (baseline) and DCNet on five test trajectories.}\label{tbl:dcnet_res}
\resizebox{0.9\textwidth}{!}{%
\begin{tabular}{lccccccc}
\hline
 & T1 $[m/s]$ & T2 $[m/s]$ & T3 $[m/s]$ & T4 $[m/s]$ & T5 $[m/s]$ & Average $[m/s]$ & Improvement $[\%]$ \\
\hline
Baseline 
& 0.74 & 0.74 & 0.74 & 0.74 & 0.74 & 0.74 & -- \\
DCNet    
& 0.21 & 0.22 & 0.21 & 0.21 & 0.26 & \textbf{0.22} & \textbf{70} \\
\hline
\end{tabular}%
}
\end{table}

\section{INS Alignment}\label{sec:ins_alg}
\counterwithout{equation}{section} 
\noindent
In underwater navigation, initial INS attitude alignment is critical for mission success, data validity, and downstream tasks such as guidance and control \cite{farrell_2022}. While roll and pitch can be accurately estimated using a stationary IMU
\cite{groves2013principles}, heading (yaw) remains more challenging, as it requires either sensing Earth’s rotation or using aiding sensors such as dual-antenna GNSS. This difficulty is further intensified under in-motion conditions, such as AUV mooring, due to the superposition of the Earth angular rate with the vehicle movements. Initial heading alignment has been studied since the 1960s \cite{Thompson_1966,Britting_1971}. Most self-alignment methods rely on vector observations in two frames, comparing IMU measurements—specific force and angular rate—with known references such as gravity and Earth’s rotation, using analytical or optimization-based solutions like Wahba’s problem \cite{Wahba_1966}. Later approaches, including the Attitude Decomposition-based Initial Alignment (ADIA) and their variantes, such as the Dual-Vector ADIA (DVA) or the Optimization-Based ADIA (OBA), integrate measurements over time \cite{Silson_2011}. However, under realistic low-dynamic or moored conditions, the observability and accuracy of DVA and OBA are limited by environmental effects, sensor noise, and short alignment periods \cite{frutuoso_2023_oceaneng}. In this section we present HeadingNet, an AI-aided framework for neural-assisted in-motion self-heading alignment.
\subsection{Model-Based ADIA approaches}
\noindent
The orientation of a given platform in time can be decomposed as follows \cite{Qin_2005}:
\begin{equation}\label{eq:general_atitude_decomp}
    \mathbf{T}_{b}^{n}(t) = \mathbf{T}_{n_0}^{n}(t)\mathbf{T}_{b_0}^{n_0}\mathbf{T}_{b}^{b_0}(t)
\end{equation}
where $\mathbf{T}_{b_0}^{n_0}$ is the constant initial alignment transformation, $\mathbf{T}_{n_0}^{n}(t)$ is the transformation of the navigation frame, and $\mathbf{T}_{b}^{b_0}(t)$ is the transformation of the body frame, both over the alignment interval $[t_0, t]$. The matrices $\mathbf{T}_{n_0}^{n}(t)$ and $\mathbf{T}_{b}^{b_0}(t)$ are computed based on the Earth’s rotation rate and the gyroscope angular rate, respectively \cite{frutuoso_2023_oceaneng}. In low-dynamics conditions, such as for a moored vessel, the gravity vector $\boldsymbol{g}^{n}(t)$ can be approximated by the specific force vector $\boldsymbol{f}^{b}(t)$ \cite{frutuoso_2023_oceaneng}. By considering the referenced frame attitude decomposition in \eqref{eq:general_atitude_decomp} the following vector observation is achieved:
\begin{equation}\label{eq:ADIA_vec_obs}
    \boldsymbol{u}^{b_0}(t) = \mathbf{T}_{n_0}^{b_0} \boldsymbol{u}^{n_0}(t)
\end{equation}
where $\boldsymbol{u}^{b_0}(t)$ is an observation vector in the body frame frozen at time $t_0$, and $\boldsymbol{u}^{n_0}(t)$ is the corresponding observation vector in the navigation frame also frozen at time $t_0$. To analytically estimate $\mathbf{T}_{n_0}^{b_0}$, DVA can be utilized where two sets of non co-linear observation vectors exist as follows \cite{Silva_2016}:
\begin{equation}\label{eq:DVA_decomp}
    \mathbf{T}_{n_0}^{b_0} = \begin{bmatrix}
        (\boldsymbol{u}^{n_0}_1)^T \\
         (\boldsymbol{u}^{n_0}_2)^T \\
          (\boldsymbol{u}^{n_0}_1 \times \boldsymbol{u}^{n_0}_2)^T
    \end{bmatrix} ^{-1}
     \begin{bmatrix}
        (\boldsymbol{u}^{b_0}_1)^T \\
         (\boldsymbol{u}^{b_0}_2)^T \\
          (\boldsymbol{u}^{b_0}_1 \times \boldsymbol{u}^{b_0}_2)^T
    \end{bmatrix}
\end{equation}
where each observation vector in both the body and navigation frames is recorded at two distinct times, $t_k$ and $t_j$ with $t_k < t_j$, yielding $\boldsymbol{u}^{n_0}_1$, $\boldsymbol{u}^{b_0}_1$ and $\boldsymbol{u}^{n_0}_2$, $\boldsymbol{u}^{b_0}_2$, respectively, thus providing two independent vector observations for DVA. In OBA, in turn, the initial attitude is estimated using Wahba’s optimal quaternion formulation \cite{Wahba_1966}. Both the DVA and OBA frameworks provide accurate alignment at longer time period, yet fail to do so in shorter time periods.
\subsection{HeadingNet Neural-assisted self-heading Alignment}
\begin{figure}[b!]
    \centering
    \includegraphics[width=0.8\linewidth]{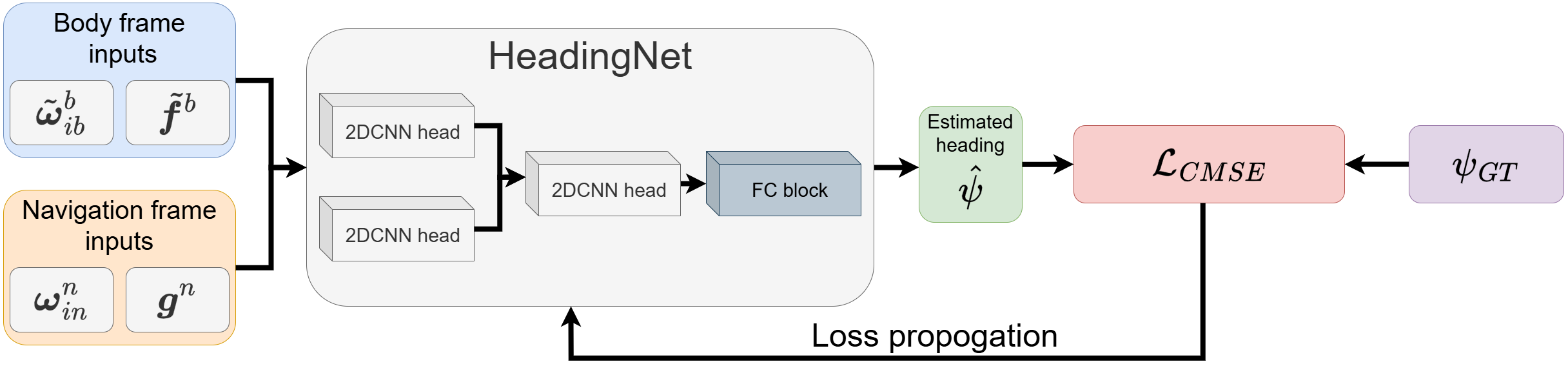}
    \caption{HeadingNet observation vector information flow and loss calculation during training.}
    \label{fig:headin_net_fig}
\end{figure}

\noindent
HeadingNet is a data-driven framework that estimates the heading angle $\psi$ at the end of the alignment period \cite{yampolsky2026neural}. This corresponds to the initial heading required by the navigation filter at mission start, rather than the heading at $t=0$. HeadingNet is a multi-head neural network architecture consisting of three 2DCNN heads followed by a fully connected (FC) block for final regression and heading estimation. For HeadingNet to function as a self-heading alignment framework, its inputs are the same as those of DVA and OBA, namely two sets of independent observation vectors: the gravity/accelerometers specific force vectors ($\boldsymbol{g}^n$, $\boldsymbol{f}^b$), and the Earth's/gyroscopes angular rates ($\boldsymbol{w}_{in}^n$, $\boldsymbol{\omega}_{ib}^b$). To train HeadingNet, a loss function adapted from \cite{ENGELSMAN2026112842} is used to account for the cyclic nature of angles, and the cyclic mean squared error (CMSE) is defined as follows:
\begin{equation}\label{eq:initial_CMSE_def}
\mathcal{L}_{\mathrm{CMSE}} = \big(\frac{1}{N}\sum_{i = 1}^{N} \boldsymbol{err}_i^2 \big) \cdot \lambda
\end{equation}
where $\lambda$ is a scaling factor, $N$ is the number of samples, and $\boldsymbol{err}_i^2$ is defined as:
\begin{equation}\label{eq:err_i}
\boldsymbol{err}_i = \mathrm{atan2}\big(\sin(\Delta \boldsymbol{\psi}_{i}), \cos(\Delta \boldsymbol{\psi}_{i})\big) \: , \forall i \in {1, \ldots,N}
\end{equation}
where $\Delta \boldsymbol{\psi}_{i}$ is the difference between the heading angle estimated by HeadingNet, $\hat{\psi}_i$, and the GT heading angle, $\psi_i$, for sample $i \in [1,\ldots,N]$. Figure \ref{fig:headin_net_fig} shows a block diagram of the information flow through HeadingNet during the training.
\\
\noindent
HeadingNet was trained and evaluated on real-world data recorded by an autonomous surface vehicle (ASV) moored to a pier on multiple days, forming a mooring-condition dataset \cite{frutuoso_2023_oceaneng}. HeadingNet was evaluated over five alignment durations: 10, 30, 60, 90, and 120 seconds allowing comparison of the alignment time efficiency and not only accuracy. Table \ref{tbl:headingnet_res} presents the averaged absolute error (AE) of the heading angle for each evaluated alignment duration for both the best model-based approach (DVA or OBA) and HeadingNet. Two main observations can be drawn from Table \ref{tbl:headingnet_res}. First, HeadingNet achieved a sub-five-degree error with only ten seconds of alignment, resulting in $60\%$ time improvement. Second, when compared to the best model-based method at 120 seconds, which yielded its highest accuracy, HeadingNet outperformed the baseline by $51\%$. 
\begin{table}[t!]
\centering
\caption{Average AE of the heading angle estimation based on the mooring-condition dataset.}
\label{tbl:headingnet_res}
\begin{tabular}{lccccc}
\hline
 & $10\,[sec]$ & $30\,[sec]$ & $60\,[sec]$ & $90\,[sec]$ & $120\,[sec]$ \\
\hline
Best model-based AE [deg] 
& 150.5 & 7.58 & 2.28 & 0.98 & 0.93 \\

HeadingNet AE [deg]       
& 4.33 & 1.53 & 1.43 & 0.97 & 0.46 \\

\textbf{Improv. [\%]} 
& \textbf{97} & \textbf{79} & \textbf{37} & \textbf{0.9} & \textbf{51} \\
\hline
\end{tabular}
\caption{Average AE of the heading angle estimation based on the mooring-condition dataset.}\label{tbl:headingnet_res}
\end{table}
\\
\noindent
To conclude, HeadingNet offers a rapid and accurate self-heading alignment which outperforms the best model-based approache. Nevertheless, HeadingNet is able to estimate only the heading angle, unlike 
DVA and OBA, 
which provide the entire initial attitude. Yet, this drawback can be tackled by expanding HeadingNet to estimate the entire orientation, or using it as a hybrid framework.
\section{INS/DVL Alignment}\label{sec:insdvl_alg}
\counterwithout{equation}{section} 
\noindent
Accurate alignment between the INS and DVL reference frames is a
prerequisite for effective sensor fusion. Due to mechanical and installation constraints, the DVL and IMU are rarely co-located or co-aligned on an AUV, introducing a constant rotation $\mathbf{C}_d^b \in SO(3)$ between their respective frames. Errors in this rotation directly corrupt velocity projection and propagate into position drift throughout the mission
\cite{kinsey2007situ, troni2010new}. In this section, we review the standard model-based approach as well an AI-aided method.
\subsection{Standard Model-Based Alignment}\label{sec:insdvl_alg_mb}
\noindent
The standard in-situ alignment approach is velocity-based, that is, the INS integrates accelerometer and gyroscope measurements to produce an estimate of inertial velocity $\boldsymbol{v}^{b}(t)$ expressed in the body frame, which is then
compared against the estimated velocity of DVL $\boldsymbol{v}^{d}(t)$. The two are
related by an unknown rotation matrix $\mathbf{T}_d^b$:
\begin{equation}\label{eq:align_vel}
    \boldsymbol{v}^{b}(t) = \mathbf{T}_d^b \, \boldsymbol{v}^{d}(t).
\end{equation}
Over $N$ time instances, the alignment is cast as a Wahba problem
\cite{wahba1965least}:
\begin{equation}\label{eq:wahba}
    \hat{\mathbf{T}}_d^b = \argmin_{\mathbf{T}_d^b \in SO(3)}
    \frac{1}{N}\sum_{i=1}^{N}
    \left\|\boldsymbol{v}^{b}[t_i] - \mathbf{T}_d^b\,\boldsymbol{v}^{d}[t_i]\right\|^2,
\end{equation}
solved by singular value decomposition (SVD) method \cite{umeyama1991least}.
This approach carries significant practical limitations. Convergence requires sufficient velocity excitation, imposing specific maneuver patterns incompatible with typical straight-line AUV operations. The method is also highly sensitive to IMU quality. For example, with navigation-grade sensors the
SVD achieves ${\sim}0.42^\circ$ RMSE, while tactical-grade sensors push this to ${\sim}23^\circ$, a tenfold degradation driven by accelerometer and gyroscope bias. Moreover, convergence times routinely exceed 70 seconds and some variants depend on external GNSS or acoustic infrastructure
\cite{zhaopeng2011online, kinsey2007situ}, limiting operational autonomy.
\subsection{ResAlignNet: Data-Driven INS/DVL Alignment}\label{sec:resalignnet}
\noindent
To overcome these limitations, ResAlignNet~\cite{damari2026resalignnet},
extending previous work on data-driven alignment~\cite{damari2025data}. It
employs a 1D ResNet-18 {\cite{he2016deep} backbone adapted for temporal sensor fusion. The alignment is reformulated as a supervised regression task that directly
estimates the three Euler angles $(\phi, \theta, \psi)$ parameterizing $\mathbf{T}_d^b$ from synchronized velocity measurements of both sensors.
The input to the network is a six-dimensional tensor
$[\mathbf{v}^b_{\mathrm{INS}}\,\|\,\mathbf{v}^d_{\mathrm{DVL}}]
\in \mathbb{R}^{W \times 6}$, where $W$ is the length of the temporal window.
An initial convolutional layer (64 filters, kernel 7, stride 2) is followed by four residual layers with channel dimensions 64, 128, 256, and 512. Each residual block implements the skip connection:
\begin{equation}\label{eq:resblock}
    \mathbf{y} = \mathcal{F}(\mathbf{x}, \{W_i\}) + \mathbf{x},
\end{equation}
which preserves gradient flow and prevents vanishing gradients
\cite{he2016deep}. A global average pooling layer then aggregates temporal features into a fixed-size representation, and a fully connected layer regresses the three alignment angles. The network minimizes the MSE between
predicted and ground-truth Euler angles:
\begin{equation}\label{eq:align_mse}
    \mathcal{L} = \frac{1}{N}\sum_{i=1}^{N}
    \sum_{j\in\{\phi,\theta,\psi\}}
    \bigl(\alpha_{i,j} - \hat{\alpha}_{i,j}\bigr)^2,
\end{equation}
using Adam optimizer (learning rate $10^{-7}$, batch size 32).
Training data are generated via a noising pipeline that applies realistic DVL and IMU error models to synthetic trajectories, enabling Sim2Real transfer. That is, a model trained on synthetic data can be deployed on operational sensor measurements without retraining. Importantly, ResAlignNet requires only onboard INS and DVL sensors, foregoing GNSS, acoustic infrastructure, and prescribed vehicle maneuvers—making it fully autonomous and eliminating pre-mission trajectory constraints.
\subsection{Experimental Results}\label{sec:resalignnet_results}
\noindent
ResAlignNet was validated using the Snapir AUV during sea trials in the Mediterranean Sea near Haifa, Israel \cite{cohen2025adaptive}. Two trajectory patterns
were evaluated: a straight-line segment (Trajectory~\#1) and a long-turn maneuver (Trajectory~\#2). Table~\ref{tab:align_summary} presents the RMSE in all window sizes and methods for both trajectories. ResAlignNet (RAN-18) consistently achieves the lowest RMSE in every window size on both trajectories. AlignNet \cite{damari2025data} offers a significant improvement over SVD but is not able to support ResNet-based architectures. ResNet-34 and CNN-LSTM reach
competitive accuracy at larger windows but show higher errors at shorter ones, while RAN-18 delivers robust sub-degree performance throughout.
The Sim2Real variant, trained exclusively on the same synthetic straight-line and turn trajectories used in simulation and then deployed directly on real-world measurements, maintains RMSE between $1.6^\circ$--$3.4^\circ$, demonstrating effective knowledge transfer from simulation to real-world deployment. The SVD baseline severely degrades under tactical-grade conditions, with maximum errors exceeding $179^\circ$ throughout the test dataset. ResAlignNet reaches accurate alignment in only 25 seconds, representing a 65\% reduction in convergence time compared to the SVD baseline \cite{damari2026resalignnet}.
\begin{table}[h!]
\centering
\caption{RMSE [$^\circ$] comparison across window sizes for Trajectories
\#1 and \#2 using real-world sea-trial data collected with the Snapir AUV.
RAN: ResAlignNet.}
\label{tab:align_summary}
\resizebox{0.75\linewidth}{!}{%
\begin{tabular}{ccccccc}
\hline
\textbf{Traj.} & \textbf{Window [s]} & \textbf{SVD} & \textbf{AlignNet} & \textbf{RAN-34} & \textbf{CNN-LSTM} & \textbf{RAN-18} \\
\hline
\multirow{5}{*}{\#1}
 & 5   & 52.47 & 1.50 & 1.02 & 1.31 & \textbf{0.95} \\
 & 25  & 14.23 & 1.25 & 0.99 & 0.92 & \textbf{0.92} \\
 & 50  & 18.47 & 1.36 & 0.94 & 0.92 & \textbf{0.91} \\
 & 75  & 24.70 & 1.32 & 0.95 & 0.92 & \textbf{0.92} \\
 & 100 & 25.74 & 1.48 & 0.94 & 0.93 & \textbf{0.91} \\
\hline
\multirow{5}{*}{\#2}
 & 5   & 59.40 & 1.49 & 1.04 & 1.14 & \textbf{0.98} \\
 & 25  & 39.64 & 1.26 & 0.97 & 0.92 & \textbf{0.93} \\
 & 50  & 44.51 & 1.24 & 0.94 & 0.92 & \textbf{0.92} \\
 & 75  & 37.44 & 1.25 & 0.91 & 0.91 & \textbf{0.90} \\
 & 100 & 36.75 & 1.45 & 0.93 & 0.91 & \textbf{0.89} \\
\hline
\end{tabular}}
\end{table}

\noindent
These results demonstrate that replacing the model-based SVD estimator with ResAlignNet yields consistent sub-degree alignment accuracy across sensor grades and trajectory types, eliminates motion-pattern requirements, and enables immediate AUV deployment without lengthy pre-mission calibration procedures.
\section{INS/DVL Fusion}\label{sec:insdvl_fus}
\counterwithout{equation}{section} 
\noindent
AUV navigation solution is primarily performed by an INS/DVL sensor fusion. It is performed by a nonlinear filtering approach that takes into consideration biases and stochastic noise. Most commonly, this is achieved using a nonlinear variant of the Kalman filter~\cite{farrell2008aided,groves2013principles}. \\
\noindent
There are two main approaches to integrating DVL measurements into the inertial dead reckoning solution to mitigate its drift. The first is the tightly coupled approach, where each raw beam measurement is processed and integrated into the filter. In this case, it does not matter how many beams are available out of the four transmitted, the filter still updates the solution to mitigate inertial drift. While fewer than three beams cannot provide accurate velocity estimates, research has shown that the tightly coupled approach can still maintain a lower divergence rate compared to using no measurements at all.
On the other hand, the loosely coupled approach first processes the raw beam measurements to estimate the Cartesian velocity vector, and then updates the filter. This velocity estimation is done using a least squares estimation. It is only feasible when three or four beams are available, if not, there is no velocity update and the navigation solution drifts~\cite{miller2010autonomous}. \\
\noindent
In both approaches, the nonlinear Kalman filter prediction and update steps are performed similarly, as described in~\cite{simon2006optimal}. The distinction lies in the measurement model. In the tightly coupled approach, each raw beam measurement is processed individually as follows:
\begin{equation}\label{eqn:dBeam}
    \boldsymbol{\delta{z}}_{i} = \boldsymbol{b}_{i}^{T}\hat{\mathbf{C}}_{n}^{b}\hat{\boldsymbol{v}}^{n}-\boldsymbol{b}_{i}^{T}\boldsymbol{v}^{b}_{DVL} 
      \approx \boldsymbol{b}_{i}^{T}\mathbf{C}_{n}^{b}\delta \boldsymbol{v}^{n} - \boldsymbol{b}_{i}^{T}\mathbf{C}_{n}^{b}\boldsymbol{v}^{n}[\times] \boldsymbol{\epsilon}
\end{equation}
where $\boldsymbol{\delta{z}}_{i}$ is the measurement residual of the $i^{th}$ beam, $\boldsymbol{b}_{i}$ is the $i^{th}$ beam unit direction vector, $\mathbf{C}_{n}^{b}$ is the rotation matrix from the navigation to the body frame, $\boldsymbol{v}^{n}$ is the velocity in the navigation frame, and $\boldsymbol{\epsilon}$ is the misalignment error vector. The corresponding measurement matrix $\mathbf{H}_{k,i}$ takes the form:
\begin{equation}\label{H_TC}
    \mathbf{H}_{k,i}^{TC}=[\boldsymbol{b}_{i}^{T}\mathbf{C}_{n}^{b} \quad -\boldsymbol{b}_{i}^{T}\mathbf{C}_{n}^{b}\boldsymbol{v}^{n}[\times] \quad \mathbf{0}_{1\times3}\quad \mathbf{0}_{1\times3}]
\end{equation}
The full $\mathbf{H}_{k}$ matrix is then assembled by concatenating the per-beam rows $\mathbf{H}_{k,i}$ according to the number of available beams.\\
\noindent
In the loosely coupled approach, a least squares estimator is first applied to the raw beam measurements to estimate the DVL velocity vector $\hat{\boldsymbol{v}}^{DVL}$, where $\boldsymbol{y}$ is the vector of beam measurements and $\mathbf{T}$ is the matrix of beam direction vectors:
\begin{equation}\label{eqn:LS}
    \hat{\boldsymbol{v}}^{DVL}=
    \underset{\boldsymbol{v}^{DVL}}{\argmin}{\mid\mid\boldsymbol{y}-\mathbf{T}\boldsymbol{v}^{DVL} \mid\mid}^{2}.
\end{equation}
The solution is obtained via the pseudo-inverse of $\mathbf{T}$ \cite{bar2001estimation}:
\begin{equation}\label{eqn:PInv}
    \hat{\boldsymbol{v}}^{DVL}=(\mathbf{T}^{T}\mathbf{T})^{-1}\mathbf{T}^{T}\boldsymbol{y}.
\end{equation}
The estimated velocity is then incorporated into the filter through the following residual:
\begin{equation}\label{eqn:LC_dz}
    \boldsymbol{\delta{z}}  = \hat{\mathbf{C}}_{n}^{b}\hat{\boldsymbol{v}}^{n}-\boldsymbol{v}^{b}_{DVL}\approx \mathbf{C}_{n}^{b}\delta \boldsymbol{v}^{n} - \mathbf{C}_{n}^{b}\boldsymbol{v}^{n}[\times] \boldsymbol{\epsilon}
\end{equation}
with the corresponding measurement matrix:
\begin{equation}\label{H_LC}
    \mathbf{H}_{k}^{LC}=[\mathbf{C}_{n}^{b} \quad -\mathbf{C}_{n}^{b}\boldsymbol{v}^{n}[\times] \quad \mathbf{0}_{3\times3}\quad \mathbf{0}_{3\times3}]
\end{equation}
\\
\noindent
The above model-based approaches are well established in the literature. However, with the rise of computational power and the proven capabilities of data-driven methods in other fields, such techniques have begun to be integrated into INS/DVL fusion methods. This is commonly achieved either by enhancing the DVL or inertial measurements prior to their incorporation into the filter, or by replacing a specific block within the filter where performance can be improved. An illustration of these approaches is shown in Fig.~\ref{fig:DL_INS/DVL}.
\begin{figure}[h!]
    \centering
    \begin{subfigure}[b]{0.49\textwidth}
        \centering
        \includegraphics[width=\textwidth]{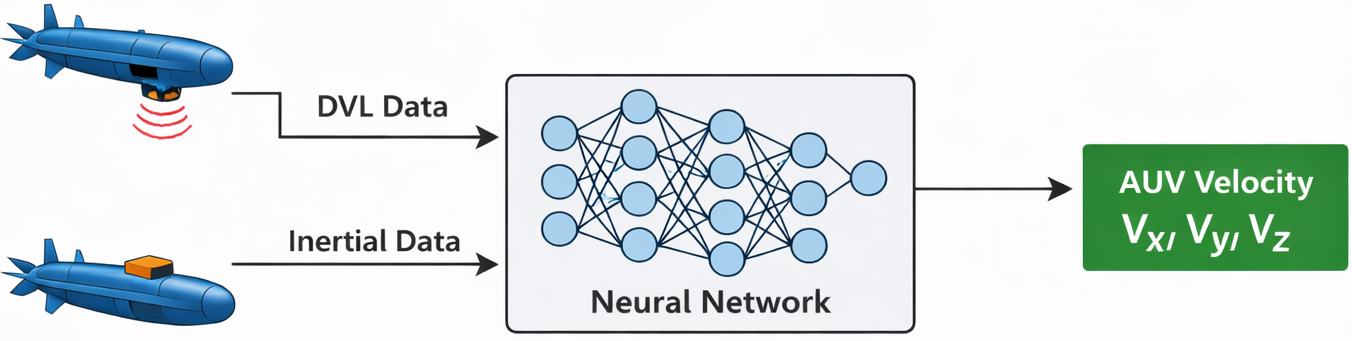}
        \vspace{3pt}
        \caption{}
        \label{fig:DL_INS/DVL1}
    \end{subfigure}
    \hfill
    \begin{subfigure}[b]{0.49\textwidth}
        \centering
        \includegraphics[width=\textwidth]{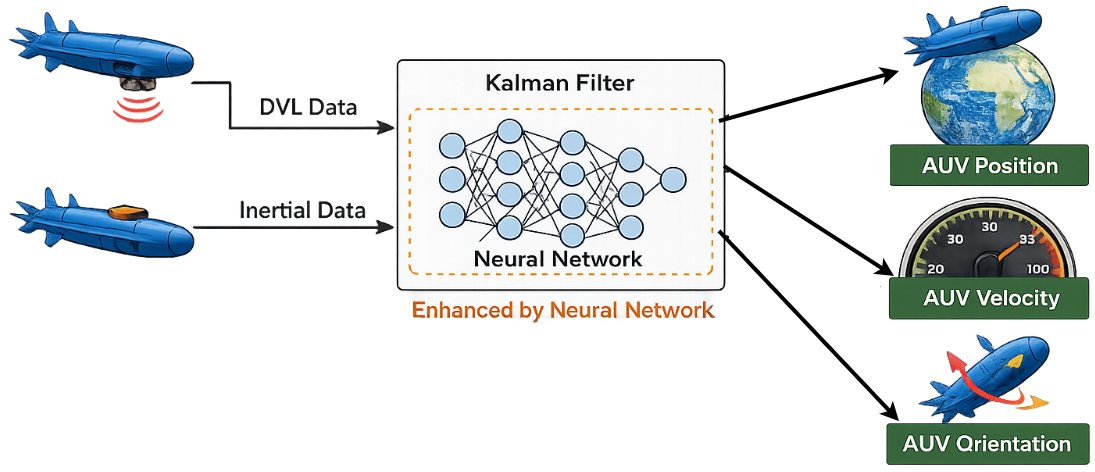}
        \caption{}
        \label{fig:DL_INS/DVL2}
    \end{subfigure}
    \caption{Illustration of the two main deep learning-based approaches to improve 
    INS/DVL navigation performance. (a) Improving DVL velocity estimation precision. 
    (b) Enhancing a single block within the Kalman filter to provide a full navigation solution.}
    \label{fig:DL_INS/DVL}
\end{figure}
\subsection{Enhancing DVL Velocity Estimation}
\noindent
Recently, a deep learning method named BeamsNet with two versions, was proposed to enhance the DVL velocity estimate~\cite{cohen2022}. The first uses 
inertial data alongside raw DVL beam measurements as input to a 
1D-CNN to regress the AUV velocity vector. The second version performs the same task using only past DVL measurements, without inertial data. In doing so, BeamsNet replaces and  enhances the least squares solution in~\eqref{eqn:PInv}. Since INS/DVL 
fusion relies on the accuracy of the DVL measurements to mitigate inertial  drift, more accurate velocity estimates directly translate to an improved  navigation solution. Both versions of BeamsNet were validated through simulation and sea experiments  conducted in the Mediterranean Sea. Results demonstrate that both versions  achieve more than 60~\% improvement in DVL velocity vector estimation accuracy  compared to the model-based least squares approach, with the first version, which  incorporates inertial data, providing the higher accuracy of the two.
Building on the capabilities of BeamsNet, it was extended to scenarios involving partial DVL measurements. The underwater environment poses extreme  conditions that may cause the acoustic beams transmitted from the DVL to  deflect away from the sensor, in scenarios such as sea creatures blocking the 
field of view, extreme roll and pitch maneuvers, uneven seabed, and more~\cite{klein2022estimating}. To this end, a hybrid neural coupled (HNC) approach was proposed in~\cite{cohen2022libeamsnet, cohen2024seamless}. For scenarios where only one or two beams are available, the BeamsNet framework was able to regress the missing beams and integrate them into the EKF via either a loosely or tightly coupled approach. The HNC approach was trained and evaluated on real AUV data recorded in the  Mediterranean Sea across two separate missions. Results demonstrate that the  proposed method outperforms the baseline loosely and tightly coupled model-based 
approaches by an average of 96.15\%, and surpasses a model-based beam estimator  by an average of 12.41\% in terms of velocity accuracy, across scenarios involving  two or three missing beams.
\subsection{Neural-Aided Adaptive Filtering}
\noindent
Another approach for utilizing deep learning within the INS/DVL fusion  framework is by enhancing a specific block within the filtering pipeline.  Specifically, in adaptive Kalman filtering, the process and/or measurement 
noise covariance matrices are estimated online throughout the mission. This  is particularly important in the underwater environment due to its extreme  and dynamic conditions. In~\cite{cohen2025adaptive}, a transformer-based  network was proposed to adapt the process noise covariance matrix in  real-time. An innovative Kalman-informed loss function was introduced,  mimicking the Kalman prediction and update steps during training, to adjust 
the process noise covariance matrix by minimizing the error of the primary  observable state, velocity, and by doing so, also maintaining the  reliability and consistency associated with Kalman filtering. The adaptive Kalman-informed transformer approach was evaluated on data recorded in the Mediterranean Sea. Results demonstrate that A-KIT outperforms the conventional EKF by more  than 49.5\% and model-based adaptive EKF approaches by an average of 35.4\%  in terms of position accuracy. It was also able to maintain filter consistency and theoretical guarantees, thereby preserving the reliability of the state estimator while improving its performance through data-driven capabilities. \\
\noindent
Considering adaptive noise estimation, two recent works integrate neural networks into the Kalman filter estimation process, demonstrating improved robustness and accuracy. In \cite{diker2026neural}, the authors propose a hybrid adaptive process noise estimation within the Lie group using a lightweight neural network architecture. The proposed hybrid approach achieves an impressive $17\%$ improvement on real-world data when trained in a sim2real paradigm. In addition, ProcessNet is proposed in {\cite{levy2026adaptive}} to adaptively estimates the process noise matrix in an unscented Kalman filter (UKF) for INS/DVL fusion. ProcessNet proposes a simple and effective end-to-end hybrid adaptive framework that outperforms both adaptive and non-adaptive baseline methods, achieving up to $34\%$ improvement over a baseline UKF.
In another work, a Gaussian process regression method was utilized to  replace the least squares solution for estimating the AUV velocity from  raw DVL beam data, while simultaneously providing an estimated covariance  matrix to be used within the EKF as an adaptive approach for the measurement 
noise covariance matrix~\cite{cohen2025gaussian}.  Results demonstrate that the approach reduces DVL velocity estimation errors by  approximately 20\% compared to the least squares approach, while simultaneously  improving overall navigation accuracy, particularly in the orientation states.  Furthermore, the adaptive covariance provided, enables a more robust 
EKF integration compared to both the LS estimator and BeamsNet.
\subsection{DVL-Based Acceleration}
\noindent
Recently, an algorithm was proposed to estimate velocity during complete DVL outages by utilizing historical DVL measurements and aiding information~\cite{klein2022estimating}. This approach also enables the estimation of the AUV’s acceleration vector, which was used as an external update to enhance INS/DVL fusion. Subsequently, a simple yet effective end-to-end deep learning approach for estimating this acceleration was introduced~\cite{11244966}. Experimental results demonstrate that this AI-aided method improves acceleration estimation by up to $67.2\%$ compared to the model-based baseline. Furthermore, the proposed approach requires only software modifications, making it highly suitable for operational AUVs.
\section{Visual–Inertial Odometry (VIO)}
\label{sec:vio}
\counterwithout{equation}{section} 
\noindent 
Underwater simultaneous localization and mapping (SLAM) aims to estimate the pose of a robotic platform while reconstructing
a representation of the surrounding environment using onboard sensing.
Unlike terrestrial and aerial systems, underwater robots operate
without access to GNSS and must rely on local sensing modalities such
as cameras, IMUs, and, in some cases,
acoustic or depth sensors. The absence of global positioning significantly increases the difficulty of long-term navigation, as
drift cannot be directly corrected through external references. \\
\noindent
Early efforts in underwater vision-based navigation have focused on monocular visual odometry due to the simplicity and low
cost of single-camera systems. Ferrera et al.~\cite{s19030687} proposed UW-VO, a monocular visual odometry framework designed
specifically for underwater environments. The method relies on optical-flow-based feature tracking and a retracking mechanism to
improve robustness under short-term occlusions and turbidity. A keyframe-based optimization with bundle adjustment is employed to
maintain trajectory consistency. While the system demonstrates improved performance compared to standard terrestrial methods, it
remains highly dependent on image quality and feature reliability. To further improve robustness in degraded underwater conditions,
Chen et al.~\cite{CHEN2025120896} proposed an adaptive image-enhanced monocular SLAM system built upon the ORB-SLAM3 framework. Their approach introduces a preprocessing stage that
combines environment recognition with adaptive image enhancement. By dynamically applying techniques such as CLAHE for low-light conditions and dark channel prior (DCP) for turbid environments,
the method significantly improves feature extraction and tracking
performance. Nevertheless, despite these improvements, monocular
systems inherently suffer from scale ambiguity and remain sensitive
to severe visual degradation. 
To overcome these limitations, stereo vision has been introduced as a more robust alternative for underwater mapping. Weidner
et al.~\cite{7989672} proposed a stereo-based framework for 3D reconstruction of underwater caves. By utilizing a dual-camera setup, the method directly recovers depth through stereo matching, thus resolving the scale ambiguity present in monocular approaches.
In addition, the authors exploit artificial illumination by detecting the intersection of the light cone with the cave boundaries, enabling robust edge-based reconstruction even under challenging lighting conditions. Visual odometry is performed using a stereo variant of
ORB-SLAM2 on well-illuminated regions, allowing estimation of the camera trajectory. Experimental results demonstrate improved
geometric consistency and reduced reconstruction noise compared to standard dense stereo methods in highly degraded underwater scenes.\\
\noindent 
Despite the advantages of stereo vision, underwater environments remain highly challenging due to dynamic illumination, repetitive textures, and motion-induced blur, which continue to
degrade feature-based tracking and data association. To address these limitations, recent works have explored event-based cameras as an alternative sensing modality for underwater SLAM. Event cameras provide high dynamic range and microsecond-level temporal resolution, making them inherently robust to rapid motion and challenging lighting conditions. \\
\noindent 
Early event-based SLAM systems, such as \cite{chen2023esvio} and \cite{niu2025esvo2}, demonstrate the potential of event-driven perception by converting asynchronous event streams into time surface (TS) representations for tracking and optimization. However, TS-based methods are highly sensitive to motion variations and texture sparsity, often leading to unstable feature extraction and matching failures, especially in degraded
underwater environments. These limitations highlight the challenges of directly applying event-based SLAM pipelines to underwater scenarios, despite their theoretical advantages. \\
\noindent 
In parallel to geometry-based approaches, recent works
have explored deep learning techniques to improve robustness of
visual odometry and SLAM in challenging underwater environments.
Teixeira et al.~\cite{teixeira2020deep} investigate the application of deep learning models for underwater visual odometry estimation, highlighting their ability to learn robust motion representations directly from image sequences. By leveraging convolutional and recurrent neural network architectures, these approaches bypass traditional feature extraction pipelines and demonstrate improved robustness under conditions of low texture, repetitive patterns, and poor illumination, which commonly degrade classical methods.
In addition, the authors propose a visual-inertial fusion network
that utilizes inertial supervision to reduce drift and improve
trajectory consistency.\\
\noindent More broadly, recent surveys have emphasized the growing role of deep learning in enhancing underwater SLAM pipelines. Heshmat et al.~\cite{heshmat2025underwater} highlight how deep learning techniques improve key components such as feature extraction, image enhancement, denoising, and sensor fusion, leading to increased robustness in low-visibility conditions. These methods also enable more reliable loop closure detection and mapping performance by leveraging data-driven representations. However, despite these advantages, deep learning-based approaches remain computationally demanding and require large-scale training data, which limits their deployment on resource-constrained underwater platforms. To further improve robustness and mitigate drift in challenging underwater environments, recent works have focused on tightly-coupled VIO systems that integrate visual measurements with inertial sensing and additional modalities. Rahman~\cite{rahman2018sonar} proposed a sonar-visual-inertial SLAM system that extends the
OKVIS framework by incorporating acoustic range measurements
from a scanning sonar. By jointly optimizing visual,
inertial, and sonar constraints in a nonlinear optimization
framework, the method improves scale estimation and robustness
in feature-degraded environments. Experimental results on
underwater caves and submerged structures demonstrate that
the inclusion of sonar significantly enhances reconstruction
accuracy and reduces tracking failures under poor visibility
conditions. \\ 
\noindent 
Building upon this multi-sensor fusion paradigm, Rahman\cite{rahman2019svin2} introduced SVIn2, a tightly-coupled SLAM system that integrates stereo vision, IMU, sonar, and depth measurements. The method incorporates robust initialization, image enhancement, and loop closure
mechanisms to address drift accumulation and localization
failures. In particular, the inclusion of depth sensing
provides additional scale constraints, while loop closure
enables global consistency over long trajectories. Results
show significant improvements in robustness and accuracy
compared to state-of-the-art visual-inertial methods,
especially in low-visibility underwater environments. More recently, Miao~\cite{miao2021univio} proposed UniVIO, a unified underwater stereo visual-inertial odometry framework that combines the advantages of feature-based and direct methods within a single optimization pipeline. The approach introduces an underwater image rectification model to compensate for refraction effects, along with a unified optimization strategy that jointly minimizes photometric and reprojection errors. This hybrid formulation improves both robustness and accuracy in
challenging conditions characterized by weak textures and
unstable illumination. Experimental evaluations demonstrate
that UniVIO outperforms existing methods across multiple
datasets, highlighting the effectiveness of combining
complementary visual representations in underwater VIO
systems.
\section{PINN-Based Dead Reckoing}\label{sec:PINN}
\counterwithout{equation}{section} 
\noindent 
Recent research has explored data-driven approaches to learn motion patterns directly from inertial data. Although these approaches demonstrate improved accuracy and generalization,  they lack interpretability, explainability and are also black-box in nature.  Alternatively, PINNs have emerged as a promising paradigm that bridges the gap between model-based and data-driven approaches. By embedding the INS equations of motion directly into the learning process, PINNs ensure that the predicted trajectories remain physically consistent while leveraging the expressive power of deep neural networks. 
MoRPI-PINN~\cite{sahoo2025morpi} and PiDR~\cite{sahoo2026pidr} are such models developed in recent times for mobile robot and AUVs, respectively.
\subsection{Physics-Informed Neural Networks for Dead Reckoning}
\noindent
Let a neural network $\mathcal{N}_\theta$ approximate the navigation states as
\begin{equation}
\hat{\mathbf{x}}(t) = \mathcal{N}_\theta(t, \mathbf{f}, \boldsymbol{\omega}),
\end{equation}
where $\hat{\mathbf{x}} = [\hat{\mathbf{p}}, \hat{\mathbf{v}}, \hat{\boldsymbol{\eta}}]$ denotes the predicted position, velocity, and orientation, respectively.
The PiDR framework is trained using a composite objective function that integrates both data-driven supervision and physics-based constraints. The data-driven loss ensures that the predicted states closely match the GT measurements by minimizing their discrepancy. It is defined as a weighted mean squared error, assigning different importance to each state component during training as in~\cite{chakraverty2025artificial}:
\begin{equation}
\mathcal{L}_{\text{data}} = \frac{1}{N} \sum_{i=1}^{N} \left( 
w_p \|\hat{\mathbf{p}}^{(i)} - \mathbf{p}^{(i)}\|_2^2 + 
w_v \|\hat{\mathbf{v}}^{(i)} - \mathbf{v}^{(i)}\|_2^2 + 
w_\eta \|\hat{\boldsymbol{\eta}}^{(i)} - \boldsymbol{\eta}^{(i)}\|_2^2 
\right),
\end{equation}
where $w_p, w_v, w_\eta$ are weighting coefficients.

\noindent
To ensure physical consistency, a physics-informed loss is introduced by enforcing the residuals of the strapdown INS equations:
\begin{equation}
\mathbf{r}_{\text{phys}} = 
\begin{bmatrix}
\frac{d\hat{\mathbf{p}}_n}{dt} - \mathbf{D}(\phi,h)\hat{\mathbf{v}}_n \\
\frac{d\hat{\mathbf{v}}_n}{dt} - \left( \mathbf{C}_n^b \mathbf{f}_b - (2\boldsymbol{\omega}_{ie}^n + \boldsymbol{\omega}_{en}^n)\times \hat{\mathbf{v}}_n + \mathbf{g}_n \right) \\
\frac{d\hat{\mathbf{C}}_n^b}{dt} - \left( \mathbf{C}_n^b \boldsymbol{\Omega}_{ib}^b - (\boldsymbol{\Omega}_{ie}^n + \boldsymbol{\Omega}_{en}^n)\mathbf{C}_n^b \right)
\end{bmatrix},
\end{equation}
and the corresponding physics loss is defined as:
\begin{equation}
\mathcal{L}_{\text{phys}} = \frac{1}{N_p} \sum_{i=1}^{N_p} \|\mathbf{r}_{\text{phys}}^{(i)}\|_2^2.
\end{equation}
The overall training objective combines both components as:
\begin{equation}
\mathcal{L}_{\text{total}} = \lambda_{\text{data}} \mathcal{L}_{\text{data}} + \lambda_{\text{phys}} \mathcal{L}_{\text{phys}},
\end{equation}
and the optimal network parameters are obtained by solving
\begin{equation} \label{eq:objective}
\theta^* = \arg\min_{\theta} \mathcal{L}_{\text{total}}(\theta).
\end{equation}
Minimizing~\ref{eq:objective} ensures that the learned trajectories are both data-consistent and physically plausible. Unlike model-based approaches, PiDR eliminates explicit numerical integration and trajectory representation via automatic differentiation. \\
\noindent
To further understand the performance of the proposed framework, we analyze the error characteristics~\cite{chakraverty2025artificial} of the predicted trajectories. Let the GT position and the predicted position be denoted as $\mathbf{p}(t)$  and $\hat{\mathbf{p}}(t)$, respectively. Then the position error is defined as:
\begin{equation}
\mathbf{e}(t) = \hat{\mathbf{p}}(t) - \mathbf{p}(t),
\end{equation}
and its magnitude is given by $ \|\mathbf{e}(t)\|_2 $. 

\noindent
Classical INS exhibits unbounded drift due to cumulative integration errors. Since position is obtained through double integration of noisy acceleration measurements, the error growth can be approximated as
\begin{equation}
\|\mathbf{e}_{\text{INS}}(t)\| \sim \mathcal{O}(t^2),
\end{equation}
which results in rapidly increasing deviation over time.

\noindent
The proposed PiDR framework demonstrates significantly improved error stability by constraining the solution space through physics-informed residuals. The learned trajectory minimizes both the data error and the physics residual, that is,
\begin{equation}
\min_{\theta} \left( \|\mathbf{e}(t)\|_2^2 + \|\mathbf{r}_{\text{phys}}(t)\|_2^2 \right),
\end{equation}
which effectively regularizes the solution. As a result, the error growth is significantly reduced and can be empirically observed to follow a bounded behavior:
\begin{equation}
\|\mathbf{e}_{\text{PiDR}}(t)\| \sim \mathcal{O}(t).
\end{equation}

\noindent These observations highlight the effectiveness of integrating physical knowledge into the learning process for drift-resilient navigation.
Although the PiDR framework estimates position, velocity, and orientation jointly, the error analysis is primarily conducted in the position domain. This is because position error inherently accumulates the effects of both velocity and orientation errors through the integration process, and thus serves as a comprehensive indicator of overall navigation performance. \\
\noindent
PiDR was evaluated on a real-world AUV dataset~\cite{cohen2025adaptive}. The dataset spans approximately 66 minutes of recorded data, including diverse motion patterns and environmental conditions as depicted in Figure~\ref{fig:pidr}. 
\begin{figure}[h!]
    \centering
    \includegraphics[width=0.5\linewidth]{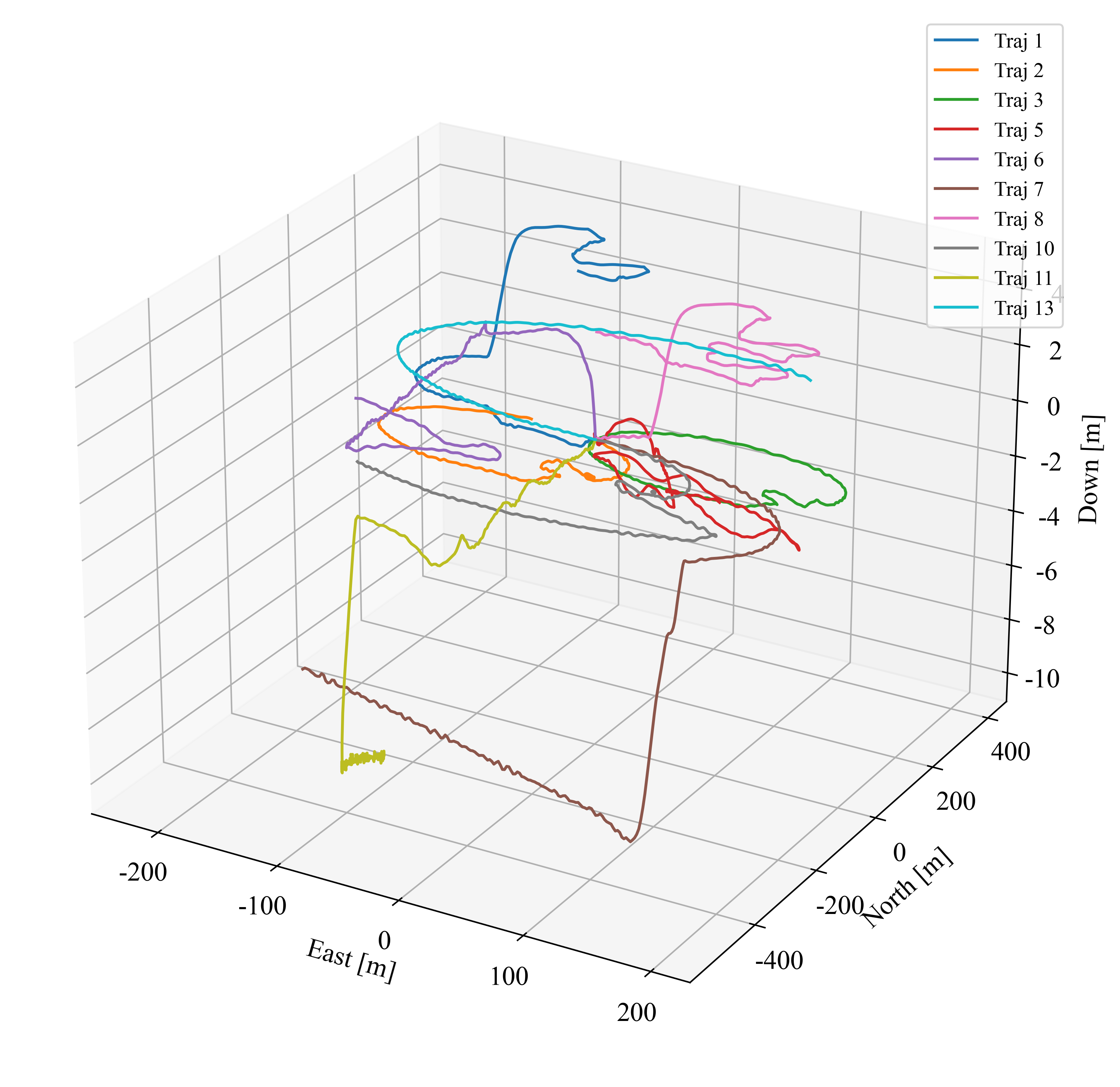}
    \caption{Trajectories T1, T2,
T3, T5, T6, T7, and T8 belong to the training set, and T10, T11,
and T13 are part of the testing set.}
    \label{fig:pidr}
\end{figure}
To evaluate the PiDR method relative to other approaches, the following metrics were used: position RMSE (PRMSE), mean absolute trajectory error (MATE), total distance error(TDR), and final distance error (FDE). From Table~\ref{tab:pidr_auv} it can be observed that PiDR achieves substantial performance gains across all metrics, with up to 97\% improvement over classical INS and consistent improvements over both MoRPI and MoRPI-PINN. This highlights the effectiveness of physics-informed learning in mitigating drift in underwater navigation.
\begin{table}[h!]
\centering
\caption{Comparative results for PiDR and baseline methods on the test dataset.}
\label{tab:pidr_auv}
\begin{tabular}{lccccc}
\hline
Method & PRMSE [m] & MATE [m] & TDE [\%] & FDE [m] & Improvement [\%] \\
\hline
3D-INS~\cite{titterton2004strapdown} & 528.3 & 392.8 & 171 & 1201.9 & 97 \\
MoRPI~\cite{etzion2023morpi} & 363.0 & 315.1 & 92 & 407.7 & 96 \\
MoRPI-PINN~\cite{sahoo2025morpi} & 256.4 & 234.2 & 81 & 297.3 & 95 \\
PiDR~\cite{sahoo2026pidr} & \textbf{14.5} & \textbf{13.0} & \textbf{5} & \textbf{12.1} & -- \\
\hline
\end{tabular}
\end{table}
\section{Conclusion}
\noindent
An accurate navigation solution is critical for the AUV’s safe return, mission validity, and overall success. Achieving this requires a pipeline of interdependent steps, where even small inaccuracies can propagate and endanger the entire mission. While the literature offers well-established model-based approaches, they rely heavily on prior knowledge and accurate modeling of the harsh, unpredictable underwater environment, which can limit their effectiveness. In DVL calibration, DCNet was shown to reduce the process complexity by offering a comprehensive error model and a simple constant velocity trajectory. Following calibration, HeadingNet exhibited neural-assisted self-alignment heading estimation, reducing the required time and increasing the accuracy significantly. ResAlignNet is later introduced for the INS/DVL alignment, showing considerable improvement over the model-based approaches, both in accuracy and time reduction. Therefore, overall the described AI-aided frameworks significantly improve accuracy and reduce the initialization time. In INS/DVL fusion, BeamsNet and AKIT were discussed, where BeamsNet enables a seamless navigation where model-based approaches fail, while AKIT offers an adaptive process noise estimation in harsh unpredictable underwater environment. In addition, SLAM-based approaches such as VIO were explored for scenarios where traditional sensors are limited or vision-based capabilities are required. Finally, PiDR, a physics-informed neural network, incorporates navigation equations as constraints within the learning process, achieving significant improvements in pure inertial navigation.\\
\noindent
The AI-aided frameworks presented in this chapter improve upon model-based methods. All were evaluated on real-world datasets collected during multiple sea missions, demonstrating robustness and practical applicability. However, these approaches have so far been tested only offline, and not in real-time. In addition, some methods, particularly SLAM-based frameworks, are computationally demanding, which can limit real-time deployment. Nevertheless, even complex SLAM models have been shown to run in real time on platforms quad-rotors, and the typically moderate underwater dynamics further support their integration into AUV navigation. Overall, improved accuracy and reduced start-up time can enhance mission reliability, reduce initialization time, and improve data quality, supporting both scientific and industrial marine applications.  

%
\section*{Acknowledgement}
\noindent
G. D. is grateful for the support of the Maurice Hatter foundation and the University of Haifa excellence scholarship for PhD studies. 
\noindent Z.Y. is supported by the Maurice Hatter Foundation and
University of Haifa presidential scholarship for outstanding
students on a direct Ph.D. track. N.C. is supported by the Maurice Hatter Foundation and University of Haifa presidential scholarship for outstanding students on a direct Ph.D. track.

\bibliography{reff}

\textbf{\textbf{\textbf{}}}\end{document}